\documentclass[conference]{IEEEtran}
\IEEEoverridecommandlockouts
\usepackage{cite}
\usepackage{amsmath,amssymb,amsfonts}
\usepackage{algorithmic}
\usepackage{graphicx}
\usepackage{textcomp}
\usepackage{xcolor}
\usepackage{hyperref}
\usepackage{booktabs}
\usepackage{multirow}
\usepackage{float}
\usepackage{listings}
\usepackage{changepage}  
\usepackage[font=small,skip=4pt]{caption}  
\usepackage{soul}

\def\BibTeX{{\rm B\kern-.05em{\sc i\kern-.025em b}\kern-.08em
    T\kern-.1667em\lower.7ex\hbox{E}\kern-.125emX}}
    
\begin{document}

\title{Optimizing Retrieval-Augmented Generation: Analysis of Hyperparameter Impact on Performance and Efficiency}


\author{%
  \IEEEauthorblockN{Adel Ammar$^{1}$, Anis Koubaa$^{2}$, Omer Nacar$^{1}$, Wadii Boulila$^{1}$}\\[2pt]
  \IEEEauthorblockA{$^{1}$Prince Sultan University,
                    Rafha Street, P.O.\ Box 66833, Riyadh 11586, Saudi Arabia\\
                    aammar@psu.edu.sa,\; onajar@psu.edu.sa,\; wboulila@psu.edu.sa}\\[4pt]
  \IEEEauthorblockA{$^{2}$Alfaisal University,
                    P.O.\ Box 50927, Riyadh 11533, Saudi Arabia\\
                    akoubaa@alfaisal.edu}%
}

\maketitle

\begin{abstract}
Large language models achieve high task performance yet often hallucinate or rely on outdated knowledge. Retrieval-augmented generation (RAG) addresses these gaps by coupling generation with external search. We analyse how hyperparameters influence speed and quality in RAG systems, covering Chroma and Faiss vector stores, chunking policies, cross-encoder re-ranking, and temperature, and we evaluate six metrics: faithfulness, answer correctness, answer relevancy, context precision, context recall, and answer similarity. Chroma processes queries 13\% faster, whereas Faiss yields higher retrieval precision, revealing a clear speed-accuracy trade-off. Naive fixed-length chunking with small windows and minimal overlap outperforms semantic segmentation while remaining the quickest option. Re-ranking provides modest gains in retrieval quality yet increases runtime by roughly a factor of 5, so its usefulness depends on latency constraints. These results help practitioners balance computational cost and accuracy when tuning RAG systems for transparent, up-to-date responses. Finally, we re-evaluate the top configurations with a corrective RAG workflow and show that their advantages persist when the model can iteratively request additional evidence. We obtain a near-perfect context precision (99\%), which demonstrates that RAG systems can achieve extremely high retrieval accuracy with the right combination of hyperparameters, with significant implications for applications where retrieval quality directly impacts downstream task performance, such as clinical decision support in healthcare.
\end{abstract}


\section{Introduction}
The emergence of Large Language Models (LLMs) represents a paradigm shift in artificial intelligence, demonstrating unprecedented capabilities in text generation, summarization, translation and complex reasoning. Despite training on vast corpora, LLMs remain prone to hallucinations—plausible yet incorrect outputs—and their static knowledge cutoff renders them unaware of developments beyond their training data. Moreover, the opaque nature of their reasoning hinders verification and accountability in high-stakes domains such as healthcare, robotics, legal analysis, and scientific research \cite{huang2023survey, zhang2023sirens, ji2023survey, ammar2024prediction, koubaa2025next}.

Retrieval-Augmented Generation (RAG) integrates external knowledge retrieval with generative models to mitigate these issues. First introduced by Lewis et al.\ \cite{RAG2020retrieval}, RAG systems retrieve semantically relevant document fragments prior to generation, grounding outputs in verifiable sources and enabling continuous updates. This dynamic interface improves factual accuracy, relevance and trustworthiness without sacrificing generative flexibility.

The adoption of RAG has expanded LLM applications in conversational AI, enterprise knowledge management and specialized professional domains. By incorporating retrieval, these systems handle complex, knowledge-intensive queries more precisely. However, retrieval incurs increased latency, complicates system architecture and raises ethical concerns around data privacy, consent and information quality—factors that demand careful design and governance.

Despite growing deployment, the impact of hyperparameter and configuration choices on RAG performance and efficiency remains underexplored. Real-world constraints—computational resources, latency requirements and accuracy thresholds—necessitate an empirical understanding of how implementation decisions affect the trade-off between responsiveness and output quality.

To fill this gap, we conducted a systematic investigation varying key components—vector stores (Chroma, Faiss), chunking strategies (naive, semantic), cross-encoder re-ranking and temperature settings—while monitoring six performance metrics: faithfulness, answer correctness, answer relevancy, context precision, context recall and answer similarity. This multidimensional analysis reveals the accuracy–latency trade-offs inherent in each choice and provides actionable insights for practitioners.

The contributions of this paper are fourfold. First, we introduce a unified evaluation framework that jointly measures RAG quality along six dimensions and tracks computational efficiency. Second, we present empirical evidence of the trade-offs associated with vector-store selection, chunking policy and re-ranking mechanisms. Third, we distil these findings into practical recommendations for balancing accuracy, responsiveness and resource utilization across application scenarios. Fourth, we extend our analysis by re-evaluating top configurations with a corrective RAG (CRAG) \cite{yan2024corrective} workflow, demonstrating that their advantages persist under iterative evidence retrieval.

The remainder of this paper is structured as follows. Section II reviews related work in RAG evaluation and optimization; Section III details our methodology, including configuration parameters, performance metrics and experimental protocol; Section IV presents experimental results; Section V discusses practical implications; and Section VI concludes with key insights and future directions.

\section{Related Work}
The development and evaluation of Retrieval-Augmented Generation (RAG) systems have evolved rapidly, with researchers proposing diverse methodologies to assess their performance across multiple dimensions. This section reviews the current landscape of RAG evaluation frameworks, efficiency optimization approaches, and the emerging understanding of configuration parameters' impact on system performance.

\subsection{Evolution of RAG Evaluation Methodologies}
Recent advances in RAG evaluation have led to increasingly sophisticated assessment frameworks that examine both retrieval and generation components. Yu et al. \cite{yu2024evaluation} introduced Auepora, a unified evaluation process that systematically examines RAG systems through three key dimensions: evaluation targets, datasets, and metrics. This comprehensive framework identifies critical assessment aspects including relevance, accuracy, faithfulness, and correctness, while also highlighting additional requirements such as latency and robustness that are often overlooked in purely quality-focused evaluations.

Several specialized benchmarks have emerged to evaluate specific aspects of RAG performance. RAGAs \cite{es2024ragas} and ARES \cite{saad2023ares} focus primarily on the relevance of retrieved documents and their impact on generation quality, providing standardized metrics for assessing information retrieval effectiveness. Similarly, RGB \cite{chen2024benchmarking} and MultiHop-RAG \cite{tang2024multihop} emphasize retrieval accuracy by comparing retrieved information against established ground truth, offering insights into the precision of document selection mechanisms. These benchmarks predominantly employ metrics that evaluate the quality of generated responses, including LLM-based judges, ROUGE, BLEU, and F1 scores to assess output accuracy against reference answers.

While these evaluation frameworks have significantly advanced our understanding of RAG quality dimensions, they typically focus on output quality metrics rather than the relationship between system configuration and performance. This emphasis on quality without corresponding attention to efficiency and hyperparameter optimization represents a significant gap in the literature that our work aims to address.

\subsection{Efficiency Optimization in RAG Systems}
The computational efficiency of RAG systems has received comparatively less attention in the research literature, despite its critical importance for practical applications. FiD-Light \cite{hofstatter2023fid} represents one of the early systematic attempts to optimize RAG systems for efficiency, highlighting execution time as a crucial factor in real-world deployments. This work demonstrated that careful architectural choices could significantly reduce computational overhead without proportional degradation in output quality.

Industry frameworks such as LangChain Benchmarks \cite{langchain2023benchmarks} and Databricks Evaluation frameworks \cite{leng2024long} have begun incorporating latency measurements as essential components of their evaluation protocols, acknowledging the practical importance of response time in user-facing applications. These frameworks provide standardized methods for measuring and comparing the computational efficiency of different RAG implementations, enabling more informed system design decisions.

The inherent trade-offs between computational efficiency and output quality in RAG systems have been increasingly recognized in recent literature. Barnett et al. \cite{barnett2024seven} identified seven critical failure points in RAG engineering, emphasizing the need for careful optimization of retrieval processes to balance response time with accuracy. Their analysis highlighted how seemingly minor configuration choices can have cascading effects on both system performance and resource utilization.

In a counterintuitive finding, Cuconasu et al. \cite{cuconasu2024power} demonstrated that introducing controlled noise into the retrieval process can sometimes improve overall system performance while potentially reducing computational overhead. This research suggests that the relationship between configuration parameters and system outcomes is often non-linear and context-dependent, underscoring the need for empirical investigation of hyperparameter effects.

\subsection{Vector Store Selection and Configuration}
Vector database selection and configuration have emerged as crucial factors affecting RAG performance. The seminal work by Johnson et al. \cite{johnson2019billion} demonstrated the effectiveness of approximate nearest neighbor search techniques like Hierarchical Navigable Small World (HNSW) and Inverted File Index (IVF) for efficient similarity search in high-dimensional spaces. These techniques form the foundation of modern vector stores used in RAG systems, enabling efficient retrieval of semantically similar documents.

Subsequent research has explored the performance characteristics of different vector database implementations under varying conditions. Comparative analyses of embedding models and vector database architectures have revealed significant performance variations depending on factors such as dataset size, query complexity, and hardware constraints. The choice between Chroma and Faiss vector stores, as examined in our work, builds upon these foundations while providing empirical evidence of their comparative advantages in different operational contexts.

Recent advances in vector indexing techniques have further expanded the configuration space for RAG systems. Innovations such as product quantization, multi-index hashing, and hybrid search approaches offer additional dimensions for optimization, each with distinct implications for retrieval accuracy and computational efficiency. These developments highlight the growing complexity of vector store configuration and the need for systematic evaluation of different approaches.

\subsection{Chunking Strategies and Their Impact}
Document chunking strategies have emerged as a critical consideration in RAG optimization. As noted by LangChain Benchmarks \cite{langchain2023benchmarks}, appropriate chunking before indexing can significantly improve retrieval effectiveness by limiting similarity computations to manageable text segments. This approach acknowledges that semantic embedding becomes less accurate for lengthy documents, and that relevant information is often contained within concise, topically coherent passages.

Research on chunking strategies has explored various approaches, from simple fixed-length segmentation to more sophisticated methods that preserve semantic coherence. Semantic chunking techniques that identify natural breakpoints in text have shown promise for improving retrieval precision, though often at the cost of increased preprocessing complexity. The trade-offs between naive chunking approaches and more sophisticated semantic methods represent an important dimension of RAG system optimization that warrants systematic investigation.

The interaction between chunking strategies and other system components adds further complexity to RAG optimization. Chunk size and overlap parameters can significantly affect both the granularity of information retrieval and the computational resources required for embedding and similarity search. These interactions highlight the multidimensional nature of RAG configuration and the need for holistic optimization approaches that consider the interplay between different system components.

\subsection{Output Quality Assessment}
The assessment of RAG system outputs has evolved beyond simple accuracy metrics to encompass more nuanced dimensions of quality. Faithfulness, which evaluates the consistency between generated content and source documents, has emerged as a critical metric for mitigating hallucinations \cite{huang2023survey}. This dimension of quality is particularly important for applications where factual accuracy and source attribution are essential requirements.

Similarly, context precision, which measures the relevance of retrieved documents to the query, provides insights into the effectiveness of the retrieval component \cite{es2024ragas, chen2024benchmarking}. This metric helps identify cases where irrelevant or tangentially related information contaminates the context provided to the language model, potentially degrading the quality of generated responses.

The development of automated evaluation methods for these quality dimensions represents an important advance in RAG assessment. LLM-based evaluation approaches that leverage the capabilities of large language models to assess output quality have gained prominence, offering scalable alternatives to human evaluation for dimensions such as relevance, coherence, and factual accuracy. These methods enable more comprehensive evaluation of RAG systems across multiple quality dimensions, though questions remain about their reliability and potential for self-preference bias \cite{panickssery2024llm}.

\subsection{Temperature Settings and RAG Performance}
Temperature settings and their impact on RAG outputs represent an area where existing research provides limited guidance. While studies on standalone LLMs have explored the effects of temperature on output diversity and accuracy \cite{achiam2023gpt}, its specific implications in the context of RAG systems—particularly regarding the balance between creativity and factual precision—warrant further investigation.

The interaction between temperature settings and retrieval mechanisms introduces additional complexity to RAG optimization. Higher temperature values may increase output diversity but potentially at the cost of faithfulness to retrieved information. Conversely, lower temperature settings may improve factual precision but could limit the model's ability to synthesize information from multiple sources. Understanding these trade-offs is essential for optimizing RAG systems for different application requirements.

\subsection{Research Gaps and Our Contribution}


Existing studies have advanced RAG evaluation and optimization but leave key gaps: joint assessment of quality and efficiency, empirical evidence of configuration trade-offs, practical deployment guidance, and understanding of iterative retrieval effects. To address these, our contributions map directly to the identified gaps. First, we introduce a unified evaluation framework that jointly measures RAG quality along six dimensions and tracks computational efficiency, filling the lack of frameworks that consider both quality and latency. Second, we present empirical evidence of the trade-offs associated with vector-store selection, chunking policy, and re-ranking mechanisms, addressing the absence of systematic data on how hyperparameter choices impact performance. Third, we distil these findings into practical recommendations for balancing accuracy, responsiveness, and resource utilization across application scenarios, supplying the deployment guidance missing from prior work. Fourth, we extend our analysis by re-evaluating top configurations with a corrective RAG (CRAG) \cite{yan2024corrective} workflow, demonstrating that their advantages persist under iterative evidence retrieval and quantifying the modest extra latency introduced.

\section{Methodology}
\begin{figure*}[t]
    \centering
    \includegraphics[width=\linewidth]{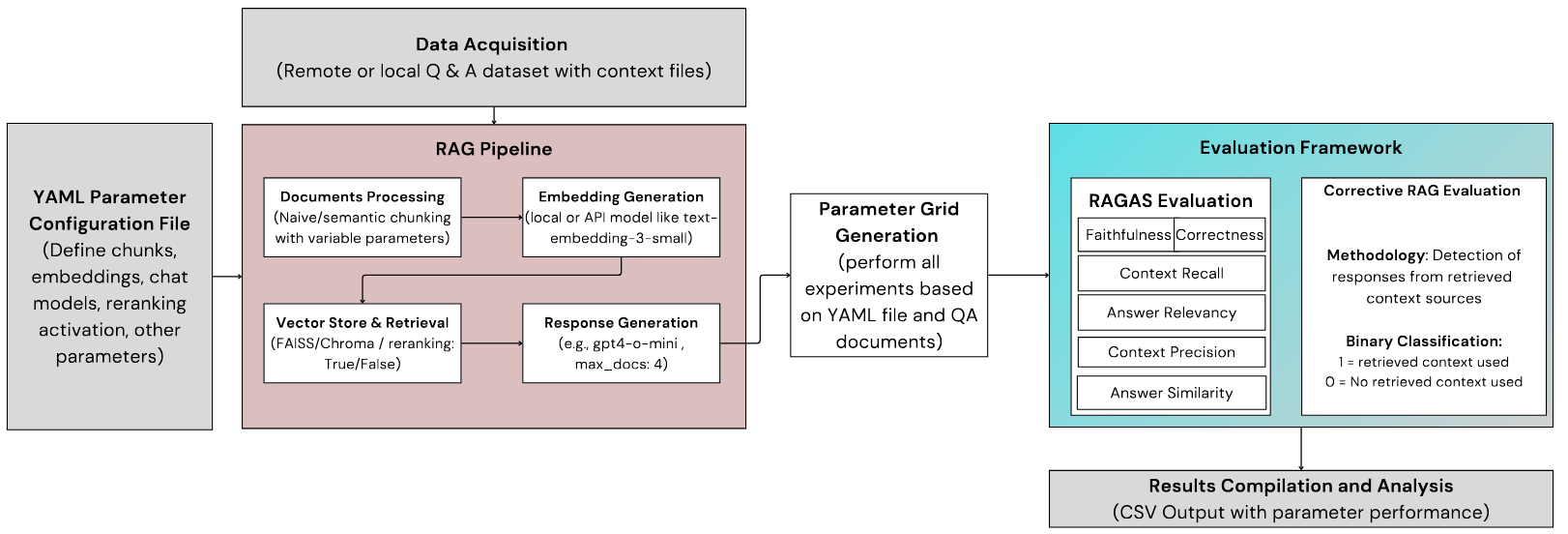}
    \caption{Systematic evaluation framework for hyperparameter optimization in RAG systems}
    \label{fig:rag-evaluation-framework}
\end{figure*}

This section presents our comprehensive framework for evaluating Retrieval-Augmented Generation (RAG) systems across multiple dimensions of performance. We designed our methodology to systematically assess how various configuration parameters affect both the quality of information retrieval and answer generation, with the goal of identifying optimal settings for different use cases and performance criteria. Figure~\ref{fig:rag-evaluation-framework} illustrates the complete evaluation workflow, from initial data acquisition through parameter grid generation, pipeline implementation, and RAGAS framework evaluation metrics.

\subsection{Evaluation Framework}
To conduct a rigorous evaluation of RAG systems, we developed a modular and configurable evaluation pipeline that enables systematic testing of different components and hyperparameters. Our framework leverages the LangChain ecosystem for implementing core RAG components and the LangSmith platform for evaluation tracking and result analysis. This architecture provides several advantages:

\begin{enumerate}
\item \textbf{Modularity}: Each component (embedding model, vector store, chunking strategy, etc.) can be independently configured and evaluated.
\item \textbf{Reproducibility}: Standardized evaluation protocols ensure consistent assessment across different configurations.
\item \textbf{Extensibility}: New components or evaluation metrics can be integrated without restructuring the entire pipeline.
\item \textbf{Traceability}: All experiments are logged with complete configuration details and performance metrics.
\end{enumerate}

The pipeline integrates document processing, vector indexing, retrieval mechanisms, and generation capabilities, all orchestrated through a YAML-based configuration system. This approach allows us to systematically explore the hyperparameter space and evaluate the impact of each configuration variable on the performance of the overall system.

For our evaluation, we utilized a question-answering dataset consisting of a diverse collection of 100 questions with short and long contexts, each paired with ground truth answers and reference contexts \cite{rag_dataset_2025}. The dataset spans multiple domains and query types, providing a comprehensive benchmark for assessing RAG system performance across various scenarios. This diversity ensures that our findings are generalizable across different application contexts and not overly specialized to particular domains or question types.

\subsection{Configuration Parameters}
Our evaluation explored a multidimensional configuration space, encompassing various components and parameters of the RAG pipeline. Table \ref{tab:experiment_setup} summarizes the key components and configuration parameters evaluated in our experiments.

\begin{table}[h!]
    \centering
    \caption{Experiment Setup and Configuration Parameters}
    \label{tab:experiment_setup}
    \begin{tabular}{|l|p{5cm}|}
        \hline
        \textbf{Component}           & \textbf{Details} \\ \hline
        Large Language Model         & \texttt{gpt-4o-mini-2024-07-18} \\ \hline
        Embedding Model              & \texttt{text-embedding-3-small} \\ \hline
        Vector Store Configurations  & Chroma, Faiss \\ \hline
        Re-ranking                   & Enabled / Disabled \\ \hline
        Maximum Tokens               & 256, 1024 \\ \hline
        Temperature                  & 0, 0.2, 0.4, 0.6, 0.8, 1 \\ \hline
        Chunking Strategy            & Naive, Semantic \\ \hline
        Chunk Size (Naive)           & 1024, 2048, 4096 \\ \hline
        Chunk Overlap (Naive)        & 128, 512 \\ \hline
        Semantic Breakpoint Thresholds & Percentile, Interquartile, Gradient, Standard Deviation \\ \hline
        Documents to Retrieve        & 2, 4 \\ \hline
        Performance Metrics          & Faithfulness, Answer Correctness, Answer Relevancy, Context Precision, Context Recall, Answer Similarity \\ \hline
    \end{tabular}
\end{table}

The primary dimensions of this configuration space include:

\subsubsection{Embedding Model}
The embedding model dimension was represented by OpenAI's \texttt{text-embedding-3-small} model, which provides an optimal balance between embedding quality and computational efficiency. This model generates 1536-dimensional vector representations of text, enabling semantic similarity search in the vector stores. We selected this model based on its strong performance in semantic similarity tasks and its widespread adoption in production RAG systems.

\subsubsection{Language Model}
For language models, we evaluated \texttt{gpt-4o-mini-2024-07-18}, a powerful yet efficient model for answer generation. This model represents a balance between computational requirements and generation capabilities, making it suitable for real-world RAG applications where both quality and efficiency are important considerations.

\subsubsection{Vector Store Implementations}
In the vector store dimension, we compared two popular implementations:

\begin{enumerate}
\item \textbf{Faiss} \cite{douze2024faiss}: Developed by Facebook AI Research, Faiss (Facebook AI Similarity Search) is known for its efficient similarity search capabilities, particularly for large-scale applications. It implements multiple indexing strategies optimized for different dataset sizes and query patterns.

\item \textbf{Chroma} \cite{xie2023brief}: A more recent vector database that offers additional features such as metadata filtering and hybrid search capabilities. Chroma provides a user-friendly API while maintaining competitive performance for moderate-sized datasets.
\end{enumerate}

Both vector stores were configured with the same embedding model to ensure a fair comparison, with default indexing parameters appropriate for the dataset size. For Faiss, we used the IndexFlatL2 index type, which provides exact nearest neighbor search. For Chroma, we used the default distance function (cosine similarity) with no additional metadata filtering.

\subsubsection{Chunking Strategies}
For document chunking, we implemented and evaluated two distinct approaches:

\begin{enumerate}
\item \textbf{Naive Chunking}: This approach splits documents into fixed-size chunks with specified overlaps, regardless of semantic content or natural document boundaries. We explored chunk sizes of 1024, 2048, and 4096 tokens with overlaps of 128 or 512 tokens.

\item \textbf{Semantic Chunking}: This more sophisticated approach identifies natural breakpoints based on semantic shifts in the text, creating chunks that preserve topical coherence. For semantic chunking, we implemented various threshold determination methods:
   \begin{itemize}
   \item \textbf{Percentile-based}: Sets breakpoints at locations where semantic shift exceeds a specified percentile of all potential breakpoints
   \item \textbf{Interquartile Range}: Uses statistical measures of dispersion to identify significant semantic shifts
   \item \textbf{Gradient-based}: Identifies points of maximum rate of change in semantic content
   \item \textbf{Standard Deviation}: Sets thresholds based on standard deviations from the mean semantic shift
   \end{itemize}
\end{enumerate}

\subsubsection{Language Model Parameters}
The language model parameters included:

\begin{itemize}
\item \textbf{Temperature Settings}: Ranging from 0.0 to 1.0 in increments of 0.2. Lower temperature values (closer to 0) produce more deterministic, focused responses, while higher values (closer to 1) introduce more randomness and creativity in the generated text.

\item \textbf{Maximum Token Limits}: We tested limits of 256 and 1024 tokens. This parameter constrains the length of generated answers, affecting both the comprehensiveness of responses and the computational resources required for generation.
\end{itemize}

\subsubsection{Retrieval Parameters}
Retrieval parameters included:

\begin{itemize}
\item \textbf{Documents to Retrieve}: We evaluated configurations retrieving either 2 or 4 documents from the vector store. This parameter affects both the amount of context provided to the language model and the computational overhead of the retrieval process.

\item \textbf{Re-ranking}: We tested configurations with and without a re-ranking step after the initial retrieval. The re-ranking mechanism uses a language model to further assess and refine the relevance of retrieved documents before answer generation, potentially improving context quality at the cost of additional computation.
\end{itemize}

This comprehensive exploration of the configuration space allowed us to identify how each parameter affects different aspects of RAG system performance and to discover optimal configurations for specific performance metrics.

\subsection{Performance Metrics}
We employed six evaluation metrics from the RAGAS framework to comprehensively assess different aspects of system quality. Each metric captures a unique dimension of RAG system performance, providing a holistic view of its effectiveness.

\subsubsection{Faithfulness Score}
The Faithfulness Score measures the factual consistency between the generated answer and the retrieved context. It penalizes hallucinations where the model produces information not supported by the context. Mathematically, it is calculated as:

\begin{equation}
\text{Faithfulness Score} = \frac{C_s}{C_t}
\end{equation}

Where:
\begin{itemize}
  \item \( C_s \) is the number of claims in the response supported by the retrieved context.
  \item \( C_t \) is the total number of claims in the response.
\end{itemize}

This metric is crucial for ensuring the reliability of the RAG system, as it directly quantifies the extent to which generated answers adhere to the information provided in the retrieved documents. A high faithfulness score indicates that the model is effectively grounding its responses in the retrieved information rather than fabricating details.

\subsubsection{Answer Correctness Score}
The Answer Correctness Score evaluates the accuracy of generated answers by comparing them against ground truth answers. It employs an F1-score calculation that balances precision and recall of factual information:

\begin{equation}
\text{F1 Score} = \frac{|\text{TP}|}{(|\text{TP}| + 0.5 \times (|\text{FP}| + |\text{FN}|))}
\end{equation}

Where TP represents true positive facts (present in both the generated answer and ground truth), FP represents false positive facts (present only in the generated answer), and FN represents false negative facts (present only in the ground truth).

This metric provides a balanced assessment of how well the generated answer captures the essential information from the ground truth while avoiding extraneous or incorrect details.

\subsubsection{Answer Relevancy Score}
The Answer Relevancy Score measures how well the generated answer addresses the specific question posed by the user. It captures whether the system stays on topic and provides pertinent information. The score is calculated as the mean cosine similarity between the original question and a set of reverse-engineered questions generated from the answer:

\begin{align}
\text{Answer Relevancy} &= \frac{1}{N} \sum_{i=1}^{N} \cos(E_{g_i}, E_o) \\
                        &= \frac{1}{N} \sum_{i=1}^{N} \frac{E_{g_i} \cdot E_o}{\|E_{g_i}\| \|E_o\|}
\end{align}

Where $E_{g_i}$ is the embedding of the $i$-th generated question, $E_o$ is the embedding of the original question, and $N$ is the number of generated questions (typically 3).

This metric helps identify cases where the system produces well-formed but off-topic responses that fail to address the user's information need, a common failure mode in generative systems.

\subsubsection{Context Precision Score}
The context Precision Score measures how effectively a retrieval system ranks relevant context items toward the top. To simplify the notation:

\begin{itemize}
  \item Let \( P@k \) be the precision at rank \( k \), defined as:
  \[
  P@k = \frac{TP_k}{TP_k + FP_k}
  \]
  where \( TP_k \) is the number of true positives at rank \( k \), and \( FP_k \) is the number of false positives at rank \( k \).
  
  \item Let \( v_k \in \{0, 1\} \) be a binary relevance indicator at rank \( k \) (1 if relevant, 0 if not).
  
  \item Let \( R_K \) be the total number of relevant items among the top \( K \) retrieved results.
\end{itemize}

Then the Context Precision@K score is defined as:
\[
CP@K = \frac{\sum_{k=1}^{K} (P@k \cdot v_k)}{R_K}
\]

Where \( K \) is the total number of top-ranked context chunks considered.

This metric is particularly important for assessing the quality of the retrieval component, as it measures how effectively the system prioritizes relevant information in the limited context window provided to the language model.

\subsubsection{Context Recall Score}
The Context Recall Score assesses how well the retrieved context covers the information required to answer the question. It measures whether the retrieval system captures all necessary information from the corpus:

\begin{equation}
\text{context recall} = \frac{|\text{GT claims that can be attributed to context}|}{|\text{Number of claims in GT}|}
\end{equation}

This metric evaluates the comprehensiveness of the retrieved information relative to the ground truth answers. A high context recall indicates that the retrieval system is successfully capturing the information needed to generate a complete and accurate response.

\subsubsection{Context Entity Recall}
The Context Entity Recall evaluates how well the entities mentioned in the ground truth answer are covered by the retrieved context. This metric provides insights into the system's ability to retrieve specific factual information:

\begin{equation}
\text{context entity recall} = \frac{|CE \cap GE|}{|GE|}
\end{equation}

Where $CE$ represents the set of entities in the context, and $GE$ represents the set of entities in the ground truth.

This metric is particularly valuable for assessing the retrieval of named entities, dates, numerical values, and other specific facts that are often critical for accurate responses in knowledge-intensive applications.

\subsubsection{Answer Similarity Score}
The Answer Similarity Score measures the semantic similarity between the generated answer and the ground truth answer. It captures whether the system produces responses that are semantically aligned with the expected answers. This metric utilizes cross-encoder models to compute semantic similarity, providing a value between 0 and 1, where higher scores indicate greater semantic alignment between the generated and ground truth answers.

These metrics were computed using the RAGAS framework's integration with LangChain, which leverages LLM-based evaluation techniques to provide a comprehensive assessment of RAG system performance across multiple dimensions.


\subsection{Experimental Protocol}
Our protocol followed a systematic grid search approach, evaluating combinations of parameters in our configuration space. This methodical exploration enabled us to identify both individual parameter effects and interaction effects between different configuration dimensions.

For each configuration, we:

\begin{enumerate}
\item Processed the document corpus using the specified chunking method and parameters.
\item Generated embeddings using the embedding model.
\item Created and configured the vector store.
\item For each question in the dataset:
   \begin{enumerate}
   \item Retrieved relevant documents using configured retrieval mechanisms.
   \item Generated an answer using the specified language model and parameters.
   \item Computed all evaluation metrics.
   \item Applied the corrective retrieval assessment.
   \end{enumerate}
\item Aggregated and analyzed results across all questions and metrics.
\end{enumerate}


To account for potential variability in network conditions and service availability, we implemented retry mechanisms with exponential backoff for all API calls and conducted experiments during periods of low service load. We also monitored and recorded execution times for each component of the pipeline, providing insights into the computational efficiency of different configurations.

This comprehensive experimental protocol enabled us to generate robust, reproducible results that illuminate the complex relationships between RAG system configuration parameters and performance outcomes across multiple dimensions of quality and efficiency.

\section{Results}
This section presents the findings from our systematic evaluation of different RAG system configurations. We analyze the impact of various hyperparameters on both performance quality and computational efficiency, highlighting key trade-offs and optimal configurations for different use cases.

\subsection{Vector Store Performance Comparison}
\begin{figure*}[!htbp]
    \centering
    \includegraphics[width=1\textwidth]{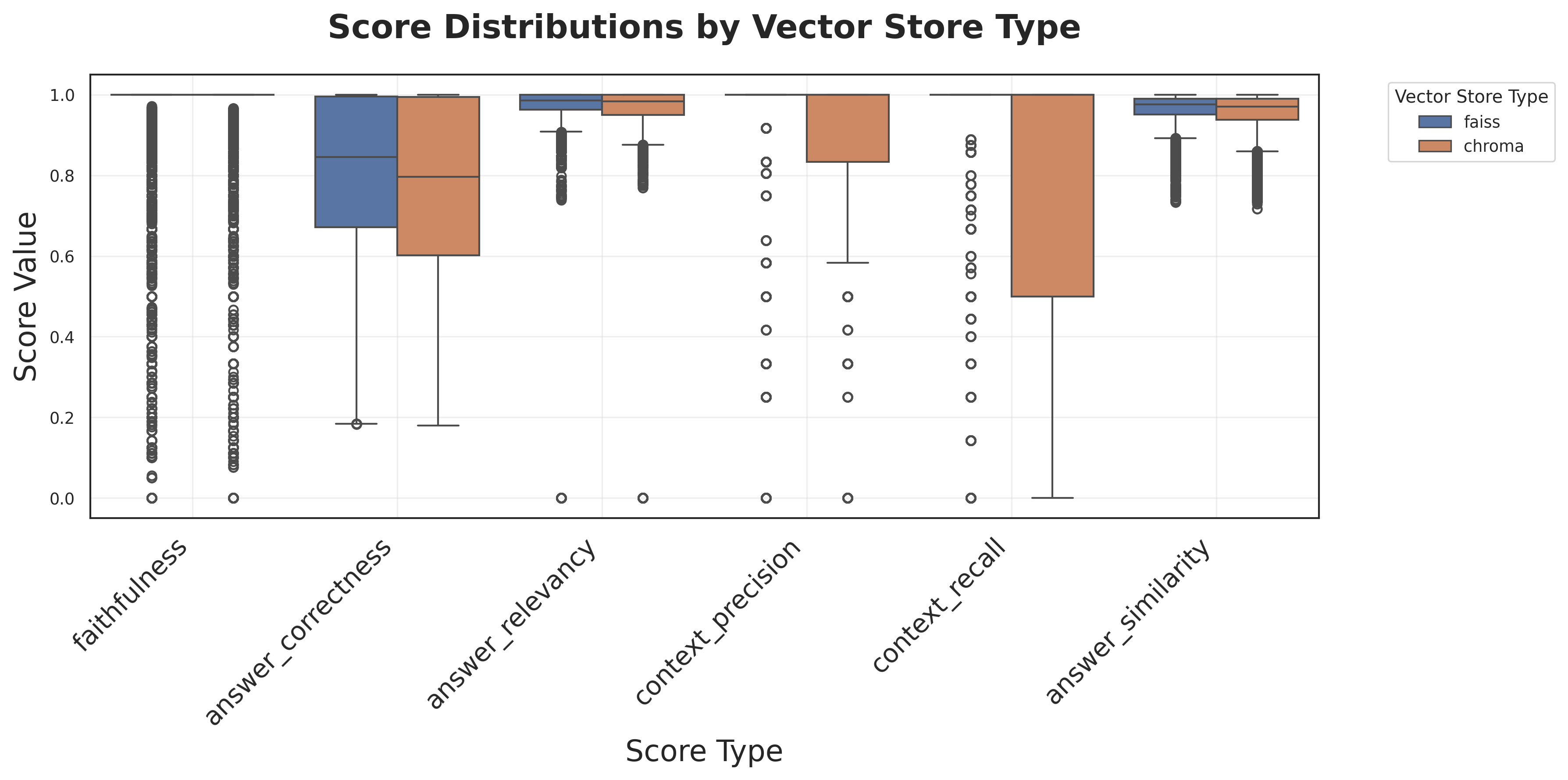}
    \caption{Score Distributions by Vector Store Type}
    \label{fig:score_distribution_by_vectorstore_type}
\end{figure*}

\begin{figure}[!htbp]
    \centering
    \includegraphics[width=0.98\linewidth]{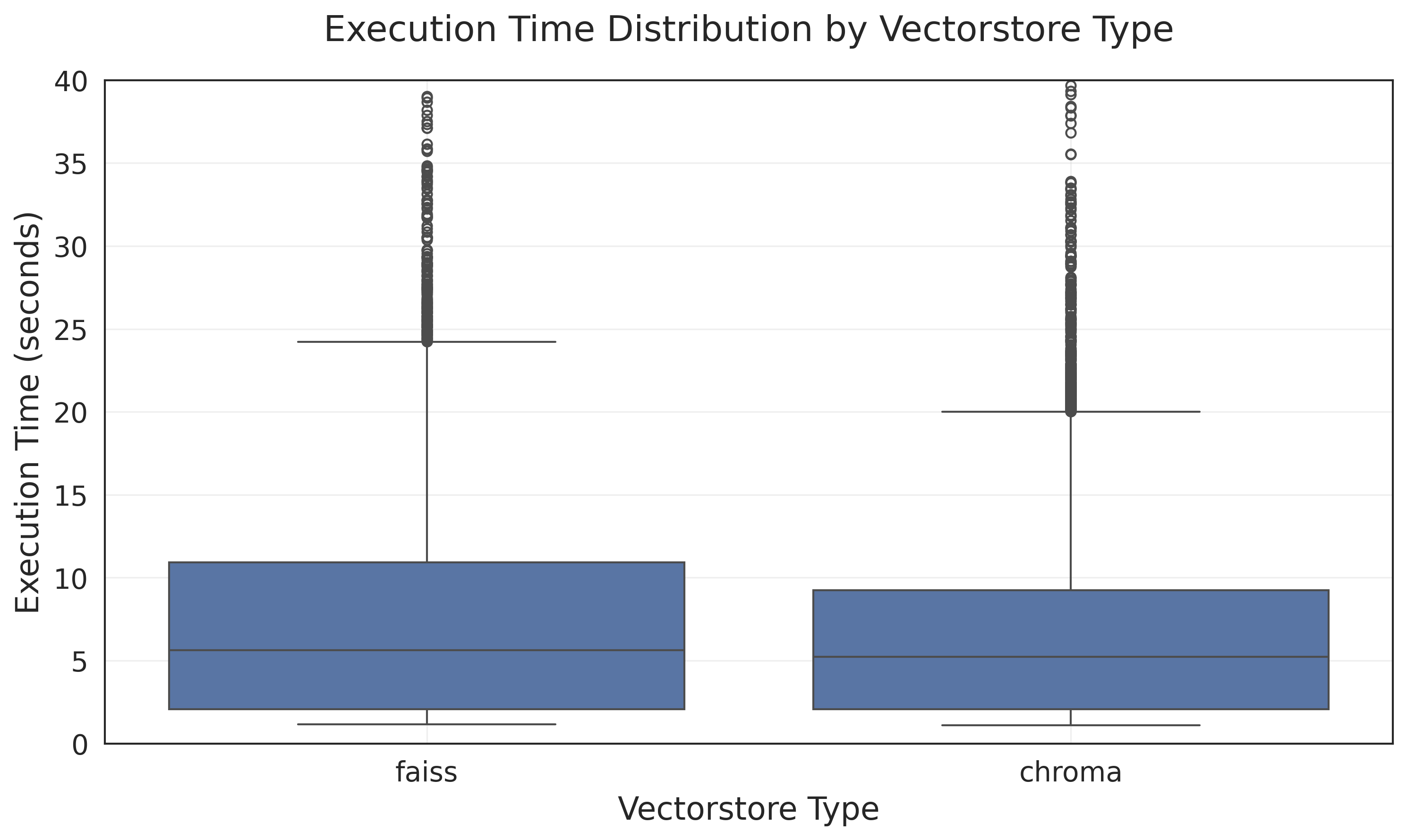}
    \caption{Execution Time Distribution by Vectorstore Type}
    \label{fig:execution_time_distribution_by_vectorstore_type}
\end{figure}

Our comparison of Chroma and Faiss vector stores revealed significant differences in both retrieval quality and computational efficiency. Figure~\ref{fig:score_distribution_by_vectorstore_type} illustrates the performance comparison across key quality metrics, while Figure~\ref{fig:execution_time_distribution_by_vectorstore_type} shows the execution time comparison.

Chroma demonstrated superior performance in terms of execution speed, with query latency approximately 13\% lower than Faiss across all test configurations. The mean query execution time for Faiss was 9.1 seconds compared to 7.9 seconds for Chroma, representing a substantial difference in real-world applications where response time is critical.

However, Faiss exhibited better retrieval quality metrics in all dimensions. The Context Precision Score for Faiss averaged 0.821 compared to 0.776 for Chroma, indicating more accurate prioritization of relevant documents. Similarly, Context Recall was approximately 6\% higher for Faiss (0.821 vs. 0.776), suggesting more comprehensive retrieval of relevant information. These quality advantages translated to downstream improvements in answer generation, with Faiss configurations achieving an average Faithfulness Score of 0.871 compared to 0.86 for Chroma.

The performance differences between vector stores varied based on other configuration parameters. For instance, when combined with semantic chunking (Table \ref{tab:semantic_chunking_scores}), the quality gap between Chroma and Faiss narrowed considerably in terms of answer correctness (0.742 vs. 0.734), answer relevance (0.853 vs. 0.866), context recall (0.762 vs. 0.748), and answer similarity (0.955 vs. 0.954), whereas it became more pronounced for faithfullness (0.861 vs. 0.782) and context precision (0.799 vs. 0.706). Conversely, with naive chunking, Faiss's quality advantage was more pronounced for all metrics ((Table \ref{tab:naive_chunking_scores})). 

\begin{table}[ht]
\centering
\small  
\caption{\parbox{.8\linewidth}{Mean evaluation scores for semantic chunking using different vector store types across six key metrics: Faithfulness, Answer Correctness, Answer Relevancy, Context Precision, Context Recall, and Answer Similarity.}}
\label{tab:semantic_chunking_scores}
\begin{tabular}{lcccccc}
\toprule
\textbf{Store} & \textbf{Faith.} & \textbf{Corr.} & \textbf{Relev.} & \textbf{Prec.} & \textbf{Recall} & \textbf{Sim.} \\
\midrule
Chroma & 0.861 & 0.742 & 0.853 & 0.799 & 0.762 & 0.955 \\
Faiss  & 0.782 & 0.734 & 0.866 & 0.706 & 0.748 & 0.954 \\
\bottomrule
\end{tabular}
\end{table}

\vspace{0.5em}

\begin{table}[ht]
\centering
\small  
\caption{\parbox{.9\linewidth}{Mean evaluation scores for naive chunking using different vector store types across the same six metrics.}}
\label{tab:naive_chunking_scores}
\begin{tabular}{lcccccc}
\toprule
\textbf{Store} & \textbf{Faith.} & \textbf{Corr.} & \textbf{Relev.} & \textbf{Prec.} & \textbf{Recall} & \textbf{Sim.} \\
\midrule
Chroma & 0.860 & 0.725 & 0.814 & 0.761 & 0.724 & 0.951 \\
Faiss  & 0.929 & 0.816 & 0.935 & 0.896 & 0.917 & 0.969 \\
\bottomrule
\end{tabular}
\end{table}


These findings suggest that the choice of vector store should be guided by application-specific priorities. For latency-sensitive applications where response time is paramount, Chroma offers clear advantages. For applications where retrieval quality and answer accuracy are more critical than speed, Faiss may be preferable. The narrowing performance gap when using semantic chunking indicates that sophisticated preprocessing can partially compensate for differences in vector store capabilities.

\subsection{Impact of Chunking Strategies}
Our evaluation of chunking strategies revealed substantial effects on both retrieval quality and computational overhead. Table \ref{tab:chunking_strategies} summarizes the performance metrics for different chunking approaches.

\begin{table}[h!]
    \centering
    \caption{Performance Comparison of Chunking Strategies. For naïve chunking, the chunk size and overlap are indicated in parentheses. For semantic chunking, the type of breakpoint threshold is specified.}
    \label{tab:chunking_strategies}
    \begin{tabular}{|l|c|c|c|c|}
        \hline
        \textbf{Chunking Strategy} & \textbf{Context} & \textbf{Context} & \textbf{Faith-} & \textbf{Execution} \\
        & \textbf{Precision} & \textbf{Recall} & \textbf{fulness} & \textbf{Time (s)} \\ \hline
Naive (1024, 128)         & 0.904 & 0.935 & 0.947 & 6.02 \\
Naive (1024, 512)         & 0.879 & 0.853 & 0.930 & 6.33 \\
Naive (2048, 128)         & 0.897 & 0.890 & 0.930 & 10.7 \\
Naive (2048, 512)         & 0.897 & 0.887 & 0.928 & 9.22 \\
Naive (4096, 128)         & 0.838 & 0.827 & 0.892 & 9.28 \\
Naive (4096, 512)         & 0.851 & 0.830 & 0.887 & 9.28 \\
Semantic (Percentile)     & 0.804 & 0.802 & 0.862 & 8.16 \\
Semantic (Interquartile)  & 0.796 & 0.793 & 0.856 & 8.65 \\
Semantic (Gradient)       & 0.790 & 0.783 & 0.843 & 8.48 \\
Semantic (Standard Deviation) & 0.621 & 0.640 & 0.727 & 9.25 \\\hline
    \end{tabular}
\end{table}

Across all evaluated quality metrics, naive fixed length chunking surpassed semantic breakpoint methods. The configuration using 1024 token windows with 128 token overlap achieved the highest precision, recall, faithfulness, and answer relevance while completing in 6.02~s, the fastest execution recorded. For the 1024 token window the larger overlap extended computation time without measurable benefit. Semantic breakpoint schemes, including percentile and inter‑quartile variants, trailed their naive counterparts on every metric and could not equal even the weakest naive setting of 4096 token windows with 512 token overlap. The superiority of fine grained fixed windows arises from three factors. Uniform coverage preserves local term proximity to query embeddings, sustaining high similarity scores, whereas semantic segmentation can generate lengthy heterogeneous sections that dilute relevant content. Modest overlaps furnish sufficient contextual look ahead for retrieval while limiting redundant tokens, thereby containing encoding and ranking cost, whereas larger overlaps mainly duplicate information. Semantic rules may also merge unrelated sentences when boilerplate or recurrent headings are misidentified, introducing noise that lowers precision. For this corpus, 1024 token fixed windows with 128 token overlap provide the optimal balance between accuracy and efficiency.

Execution time for naïve chunking decreased as window size increased, because larger windows reduce the total number of embedding calls. The sole deviation from this trend—observed for the 2048‑token window with 128‑token overlap, whose runtime exceeded that of the 1024‑token setting—is best ascribed to ordinary stochastic variability in the execution environment, such as transient contention for compute resources, cache effects, or thread‑scheduling fluctuations, rather than to any systematic property of the configuration itself.

\subsection{Re-ranking Effects}
\begin{figure*}[!htbp]
    \centering
    \includegraphics[width=1\textwidth]{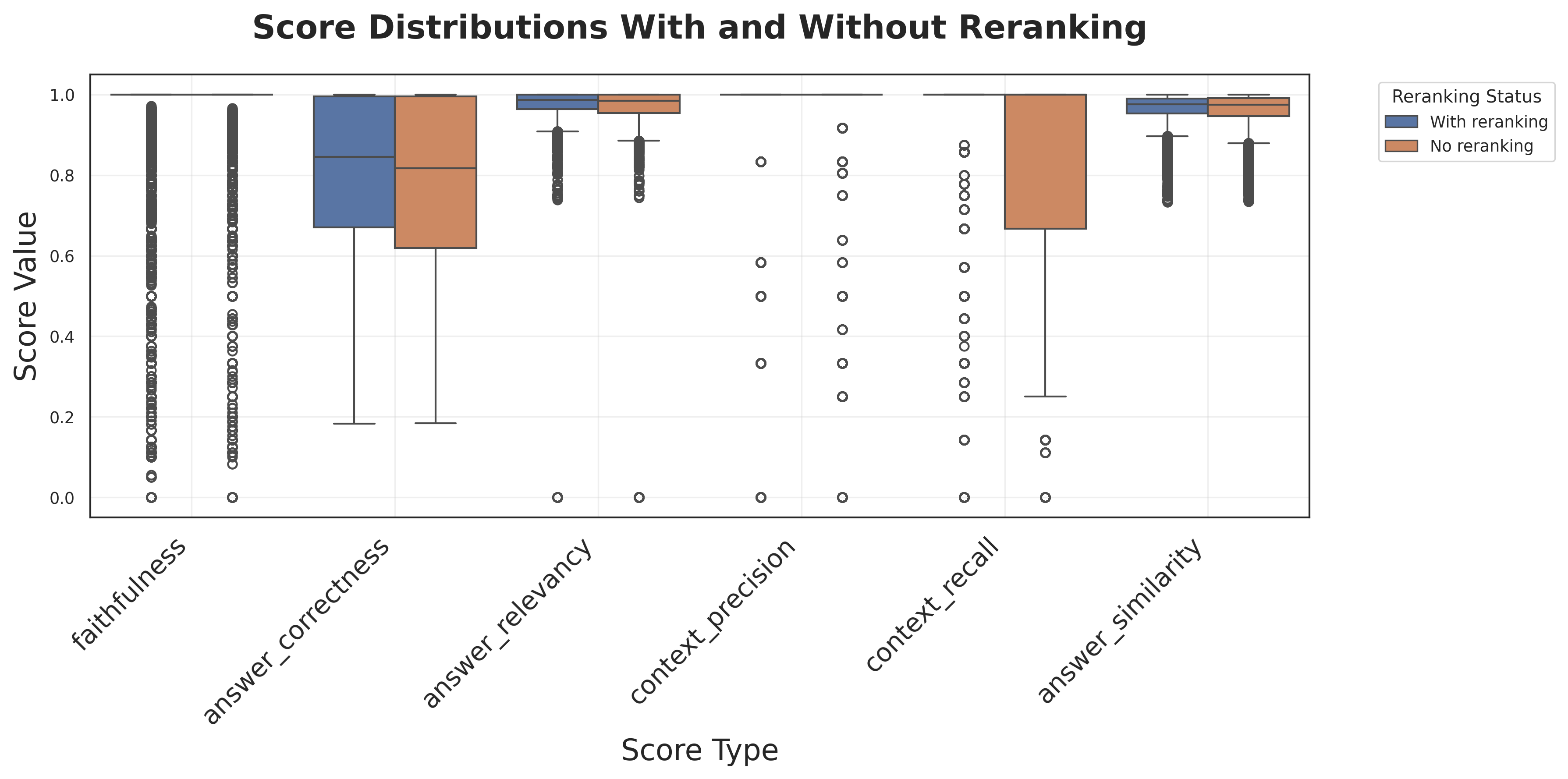}
    \caption{Score Distributions With and Without Reranking}
    \label{fig:score_distribution_w_wo_reranking}
\end{figure*}

The inclusion of a re-ranking step in the retrieval pipeline had a substantial impact on both retrieval quality and computational overhead. Figure~\ref{fig:score_distribution_w_wo_reranking} illustrates the performance differences between configurations with and without re-ranking.

Introducing a re‑ranking stage raised retrieval effectiveness for every configuration tested: mean Context Precision climbed from 0.80 to 0.85 (6\% relative gain) and Context Recall from 0.78 to 0.86 (10\%). The richer evidence base, however, translated only weakly to answer‑generation quality, with Faithfulness rising by 0.6\% and Answer Correctness by 4\%. This muted downstream response indicates that the generator had already reached a point of diminishing returns with respect to additional relevant context; most residual errors stem from reasoning or synthesis limitations in the language model rather than from gaps in the retrieved material.

The gains delivered by re-ranking were offset by a sharp rise in compute demand: introducing the second pass multiplied total runtime by roughly five in every configuration because each query document pair had to be evaluated again with a more expensive cross encoder. This additional scoring phase dominated the overall pipeline, dwarfing the modest overhead of simple retrieval and prompting. In practical deployments such a cost increase may be hard to justify, since the corresponding improvements in faithfulness and answer correctness were marginal. The result highlights a diminishing returns effect: once the primary retriever supplies a reasonably accurate context, further refinements through re-ranking provide only incremental benefit while imposing substantial latency and resource usage.

These findings suggest that re-ranking should be selectively applied based on application requirements. For use cases where accuracy is paramount and latency is less critical, such as research applications or legal document analysis, re-ranking provides valuable quality improvements that justify the computational overhead. For applications with strict latency requirements, such as real-time customer support or interactive search, re-ranking might be reserved for complex queries or cases where initial retrieval quality appears insufficient.

\subsection{Analysis of Parameter Impact on Execution Time}

\begin{figure*}[!htbp]
    \centering
    \includegraphics[width=0.8\textwidth]{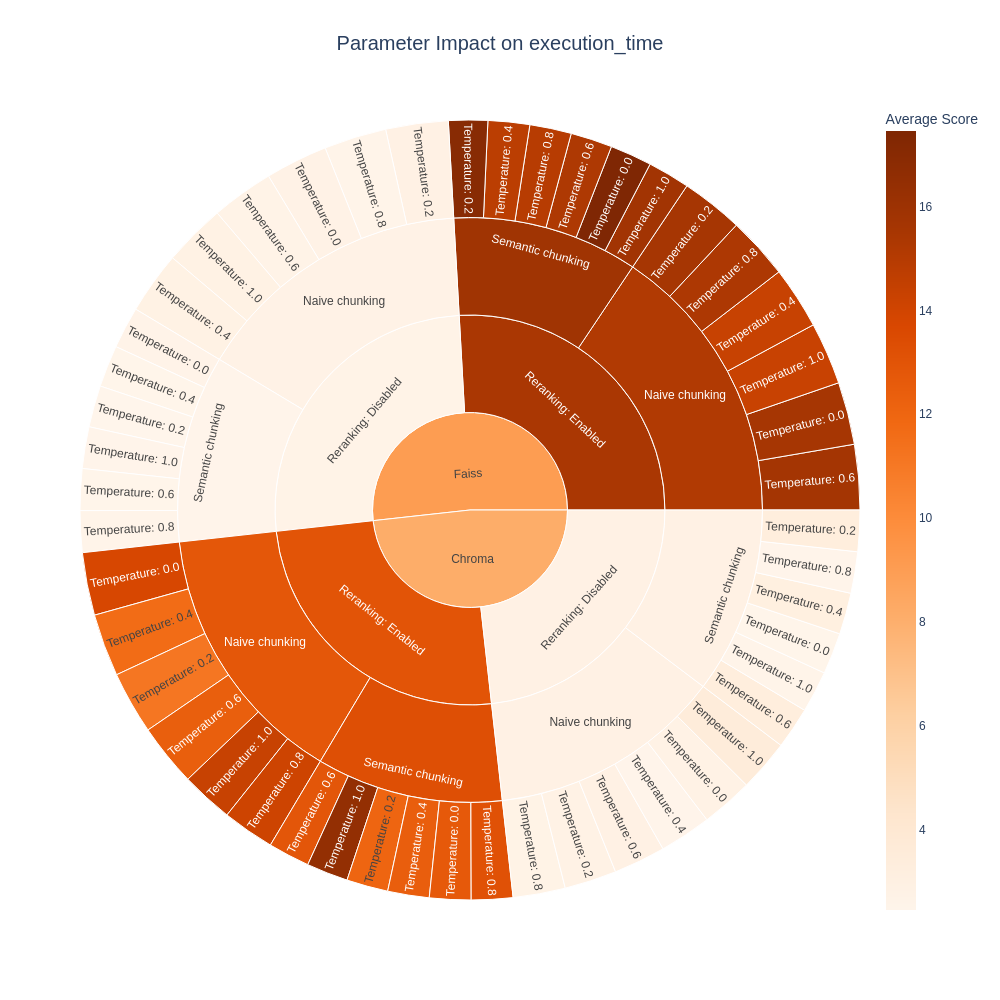}
    \caption{Parameter Impact on Execution Time}
    \label{fig:sunburst_execution_time}
\end{figure*}

The sunburst chart in Figure \ref{fig:sunburst_execution_time} offers a nuanced view of how hyperparameters affect the execution time of a RAG model, revealing intricate interdependencies. The core components, \textbf{Faiss} and \textbf{Chroma}, serve as the foundation, with Chroma exhibiting a broader execution time range, suggesting its heightened sensitivity to parameter variations. This sensitivity could be attributed to Chroma's architecture, which might be more adaptable or complex, thus reacting more significantly to parameter changes. 

The impact of \textbf{re-ranking} is pronounced, as enabling it consistently increases execution time, aligning with the expectation that re-ranking introduces additional computational overhead. This finding underscores the need for balancing computational cost with potential gains in model accuracy. The choice between \textbf{naive} and \textbf{semantic chunking} further complicates execution time dynamics. Semantic chunking, particularly when paired with re-ranking, demands more processing power, likely due to its sophisticated nature, which involves deeper semantic understanding. 


Additionally, certain external factors can contribute to time delays during execution. These include:
\begin{itemize}
    \item \textbf{Temporary network issues}, which can disrupt data transfers and slow down model responses.
    \item \textbf{Rate limiting}, where service providers limit the number of requests that can be processed within a given time frame.
    \item \textbf{Service temporary unavailability}, which may cause retries and increase execution duration.
    \item \textbf{Transient errors}, which require additional time for error handling and recovery procedures.
\end{itemize}

These factors, combined with the internal dynamics of hyperparameter settings, emphasize the necessity of strategic hyperparameter selection, balancing execution efficiency with model performance. Optimizing for both computational costs and model accuracy is crucial in real-world applications where resources are often constrained.

\subsection{Evaluating Configurations Using Corrective RAG}
To determine whether the gains reported in Sections IV-A to IV-C persist when the generator is granted an opportunity to query for missing evidence, we re-examined the top-ranked settings under a corrective retrieval-augmented generation (CRAG) \cite{yan2024corrective} workflow. For each of the six evaluation dimensions—faithfulness, answer correctness, answer relevancy, context precision, context recall, and answer similarity—we selected the configuration that achieved the highest mean score in the baseline grid search (covering vector store choice, chunking parameters, and re-ranking). These six “best-of-metric” variants were then re-executed with CRAG, in which the language model iteratively requests additional documents whenever its draft answer lacks sufficient textual support. The analysis that follows compares baseline and CRAG outcomes to quantify the added value of corrective retrieval for configurations that already occupy local optima in the conventional RAG space, and to reveal any associated latency penalties.

The corrective retrieval assessment revealed significant variation in retrieval relevance across configurations. Table~\ref{tab:retrieval_results} presents the performance of each configuration in terms of retrieval relevance.


\begin{table*}[htbp]
    \centering
    \caption{Retrieval Relevance by Configuration}
    \begin{tabular}{|l|c|c|c|c|c|c|c|c|}
        \hline
        \textbf{Configuration} & \textbf{Reranking} & \textbf{Vector Store} & \textbf{\begin{tabular}[c]{@{}c@{}}Max\\ Tokens\end{tabular}} & \textbf{Temp.} & \textbf{\begin{tabular}[c]{@{}c@{}}Max\\ Docs\end{tabular}} & \textbf{\begin{tabular}[c]{@{}c@{}}Chunking\\ Method\end{tabular}} & \textbf{\begin{tabular}[c]{@{}c@{}}Chunk\\ Overlap\end{tabular}} & \textbf{Score} \\ \hline
        Faithfulness Score      & False & Faiss   & 1024 & 0.4 & 4 & naive & 128 & 85\% \\ \hline
        Answer Correctness Score & False & Faiss   & 1024 & 0.0 & 4 & naive & 128 & 87\% \\ \hline
        Answer Relevancy Score   & False & Faiss   & 1024 & 0.2 & 4 & naive & 128 & 88\% \\ \hline
        Context Precision Score  & True  & Chroma & 1024 & 0.4 & 4 & naive & 512 & 99\% \\ \hline
        Context Recall Score     & False & Faiss   & 1024 & 0.8 & 4 & naive & 128 & 86\% \\ \hline
        Answer Similarity Score  & False & Faiss   & 1024 & 1.0 & 4 & naive & 128 & 86\% \\ \hline
    \end{tabular}
    \label{tab:retrieval_results}
\end{table*}

As shown in~\ref{tab:retrieval_results}, the Context Precision Score configuration demonstrated exceptional performance, achieving a 99\% relevance rate. This configuration uniquely employs a reranking step and uses Chroma as its vector store, with a substantially larger chunk overlap (512.0) compared to other configurations.

Among the non-reranking configurations that use FAISS, the Answer Relevancy Score configuration performed best with an 88\% relevance rate, likely due to its balanced temperature setting (0.2) that provides controlled randomness in retrieval.

The dramatic performance improvement observed in the Context Precision Score configuration (99\% relevance) strongly suggests that reranking plays a crucial role in filtering out irrelevant documents. This additional step appears to justify the computational overhead it introduces.

While most configurations utilized FAISS, the highest-performing configuration employed Chroma. This suggests that different vector databases may offer inherent advantages depending on the query types or document collections.The Context Precision Score configuration utilized a significantly larger chunk overlap (512.0 vs 128.0). This may help prevent important context from being split across chunks, leading to more coherent and relevant retrievals.

Configurations with moderate temperature settings (0.2--0.4) generally outperformed those with extreme settings (0.0 or 1.0). This suggests that a controlled degree of randomness enhances retrieval diversity without sacrificing relevance. The near-perfect performance of the Context Precision Score configuration (99\% relevance) demonstrates that RAG systems can achieve extremely high retrieval accuracy with the right combination of hyperparameters. This finding has significant implications for applications where retrieval quality directly impacts downstream task performance.

\subsection{Implications for RAG System Design}

The near-perfect performance of the Context Precision Score configuration (99\% relevance) demonstrates that RAG systems can achieve extremely high retrieval accuracy with the right combination of hyperparameters. This finding has significant implications for applications where retrieval quality directly impacts downstream task performance.

Our results emphasize the importance of comprehensive hyperparameter optimization in RAG system design. The substantial performance variations observed across configurations highlight that default settings may significantly underperform compared to optimized configurations. This underscores the value of systematic evaluation frameworks like the one presented in this study for identifying optimal configurations for specific use cases.

Furthermore, our findings suggest that different optimization targets (such as faithfulness, answer correctness, or context precision) may require distinct configuration approaches. This indicates that RAG system designers should consider their specific performance priorities when selecting and tuning system components.

\section{Discussion}
This section interprets our experimental findings in a broader context, examining the implications for RAG system design and implementation. We analyze key trade-offs between performance and efficiency, provide practical recommendations for different use cases, and acknowledge limitations of our study.

\subsection{Performance-Efficiency Trade-offs}
Our results reveal several fundamental trade-offs in RAG system design that practitioners must navigate when implementing these systems in real-world applications. These trade-offs represent critical decision points where improvements in one dimension often come at the expense of another.

The most prominent trade-off exists between retrieval quality and computational efficiency. The configurations that achieved the highest metrics, notably those combining a re ranking stage with the Faiss vector store, also recorded the longest runtimes because the cross‑encoder pass and additional distance calculations enlarged the query workload. This trade‑off varies across pipeline components. On the other hand, in the chunking stage, naïve fixed‑length windows with small size and minimal overlap boosted quality scores while leaving processing time almost unchanged, showing that careful segmentation can improve accuracy without a proportional rise in cost.

These multidimensional trade-offs highlight the importance of context-specific optimization in RAG system design. Rather than seeking a universally optimal configuration, practitioners should calibrate system parameters based on the specific requirements, constraints, and priorities of their application domain.

\subsection{Practical Implications}
Our findings have several practical implications for implementing RAG systems across different domains and use cases. We offer the following recommendations based on our experimental results:

\subsubsection{Domain-Specific Optimization Strategies}
Different application domains have distinct requirements that should guide configuration choices:

\begin{enumerate}
\item \textbf{Legal and Medical Applications}: These domains require high factual precision and source fidelity. We recommend configurations prioritizing Faithfulness and Context Precision: naive chunking, re-ranking enabled, Faiss vector store, and low temperature settings (0.0-0.2). Despite the computational overhead, the accuracy benefits justify the additional latency in these high-stakes domains.

\item \textbf{Customer Support and Real-Time Applications}: These use cases demand rapid response times while maintaining reasonable accuracy. We recommend more balanced configurations: naive chunking with moderate chunk size (1024 tokens) and overlap (128 tokens), Chroma vector store, selective re-ranking for complex queries only, and moderate temperature settings (0.2-0.4). This approach reduces average latency while preserving acceptable quality metrics.

\item \textbf{Educational and Content Creation Applications}: These domains benefit from engaging, diverse outputs while maintaining factual grounding. We recommend configurations that balance creativity and accuracy: naive chunking, re-ranking enabled, either vector store depending on scale requirements, and moderate temperature settings (0.4-0.6). This approach produces more natural, varied responses while maintaining reasonable faithfulness to source material.

\item \textbf{Research and Analytical Applications}: These use cases involve complex information synthesis across diverse sources. We recommend configurations maximizing retrieval comprehensiveness: naive chunking, re-ranking enabled, Faiss vector store with higher document retrieval counts, and low to moderate temperature settings (0.2-0.4). This approach optimizes for context recall and answer correctness at the expense of latency.
\end{enumerate}

\subsubsection{Robustness Enhancements}
Based on our analysis of external factors, we recommend several approaches to enhance RAG system robustness:

\begin{enumerate}
\item \textbf{Graceful Degradation}: Implement tiered fallback mechanisms that progressively simplify the retrieval process under degraded conditions. For instance, if re-ranking fails or experiences high latency, the system can fall back to the initial retrieval results rather than failing completely.

\item \textbf{Caching Strategies}: Implement multi-level caching for embeddings, retrieval results, and even generated answers for common queries. This can significantly reduce the impact of network issues and service disruptions on overall system performance.

\item \textbf{Asynchronous Processing}: For non-interactive applications, consider asynchronous processing patterns that decouple retrieval and generation steps. This approach can improve throughput and resilience at the cost of increased end-to-end latency for individual queries.

\item \textbf{Monitoring and Adaptation}: Implement comprehensive monitoring of both performance metrics and system health indicators. Adaptive systems that can adjust configuration parameters based on current conditions (e.g., reducing retrieval complexity during periods of high load) demonstrate better overall reliability.
\end{enumerate}

These practical recommendations provide a starting point for domain-specific RAG implementation. However, we emphasize that optimal configurations should ultimately be determined through empirical testing with domain-specific data and use cases.

\subsection{Limitations}
While our study provides valuable insights into RAG system optimization, several limitations should be acknowledged:

\subsubsection{Dataset Limitations}
Our evaluation dataset, while diverse, may not fully represent the complexity and specificity of all real-world applications. The 100 questions span multiple domains and query types, but domain-specific applications may encounter more specialized query patterns or require different optimization priorities than those identified in our general-purpose evaluation.

Additionally, our dataset focuses primarily on factual question-answering tasks. Other RAG applications, such as creative writing assistance, code generation, or conversational agents, may exhibit different sensitivity to configuration parameters and require alternative optimization strategies.

\subsubsection{Generalizability Constraints}
Our experiments were conducted with specific models (\texttt{gpt-4o-mini-2024-07-18} for generation, \texttt{text-embedding-3-small} for embeddings) and may not generalize perfectly to other model architectures or sizes. Different language models may exhibit varying sensitivity to retrieval quality, context length, and temperature settings.

Furthermore, our evaluation focused on English-language content. Multilingual RAG systems may face additional challenges related to embedding quality, chunking effectiveness, and retrieval precision that are not captured in our monolingual evaluation.

\subsubsection{Technical Constraints}
Our evaluation framework has several technical limitations:

\begin{enumerate}
\item \textbf{Metric Limitations}: While we employed a comprehensive set of evaluation metrics, automated evaluation of aspects like factual accuracy and relevance remains challenging. LLM-based evaluation approaches, while powerful, may exhibit biases or inconsistencies that affect metric reliability.

\item \textbf{Hyperparameter Space}: Despite our systematic exploration of the configuration space, we could only evaluate a finite subset of possible parameter combinations. The high dimensionality of the hyperparameter space means that potentially optimal configurations may exist outside our tested parameter ranges.

\item \textbf{Long-term Performance}: Our evaluation focused on immediate performance metrics and did not assess long-term aspects such as index maintenance requirements, embedding drift over time, or adaptation to evolving document collections.
\end{enumerate}

These limitations suggest several directions for future research, including domain-specific optimization studies, evaluation of emerging model architectures, exploration of adaptive configuration approaches, and development of more robust evaluation methodologies for RAG systems.

\subsubsection{Ethical and Privacy Limitations}
Our evaluation does not address some ethical and privacy considerations that arise in real‐world RAG deployments:
\begin{enumerate}
  \item \textbf{Privacy Leakage}: RAG systems may inadvertently retrieve and surface sensitive or personally identifiable information from user queries or document collections. Without privacy‐preserving mechanisms (e.g., differential privacy or token filtering), this can violate data protection requirements and user trust.
  \item \textbf{Consent and Data Governance}: We did not evaluate mechanisms for obtaining explicit user consent or enforcing governance policies over which documents may be indexed and retrieved. In regulated domains (e.g., healthcare, finance), such omissions can render a system non‐compliant with standards like GDPR or HIPAA.
  \item \textbf{Bias and Fairness}: Retrieval and generation components may amplify biases present in the underlying corpora or embedding models. Our monolingual and general‐purpose dataset does not permit analysis of disparate impacts across demographic groups or content types.
  \item \textbf{Security of Embeddings and Indices}: Embedding vectors and vector store indices can be vulnerable to extraction or inversion attacks, allowing adversaries to reconstruct sensitive training data. Our framework does not include threat modeling or secure storage protocols.
\end{enumerate}

\section{Conclusion}
This study has provided a comprehensive analysis of how various hyperparameters and configuration choices affect the performance and efficiency of Retrieval-Augmented Generation (RAG) systems. Through systematic experimentation with different vector stores, chunking strategies, re-ranking mechanisms, and temperature settings, we have identified key trade-offs and optimization opportunities that can guide the implementation of RAG systems across diverse application domains.

Our findings demonstrate that RAG system performance is highly sensitive to configuration choices, with significant variations in both output quality and computational efficiency depending on the selected parameters. Vector store selection presents a clear trade-off between retrieval quality and query latency, with Faiss offering superior retrieval precision at the cost of increased execution time compared to Chroma. Chunking strategies exhibit a different quality-efficiency balance, with naive chunking methods providing both substantially better retrieval quality and less computational resources than semantic approaches. The inclusion of re-ranking in the retrieval pipeline consistently improved output quality across all configurations. However, these improvements came at a considerable computational cost, increasing overall execution time by 5 times. 

Beyond these component-specific effects, our research revealed important interactions between different configuration parameters. The benefits of sophisticated components like re-ranking were more pronounced when combined with simpler chunking strategies, suggesting that targeted optimization of specific pipeline components may be more efficient than maximizing quality across all dimensions simultaneously. 

These insights have important implications for RAG system implementation in real-world applications. For domains where factual precision is paramount, such as legal or medical applications, configurations prioritizing retrieval quality (re-ranking, low temperature) are clearly preferable despite their computational overhead. For latency-sensitive applications like customer support, more balanced configurations (selective re-ranking, Chroma vector store) offer a better compromise between quality and efficiency. Educational and content creation applications may benefit from configurations that balance creativity and accuracy through moderate temperature settings and high-quality retrieval mechanisms.

The significance of this work extends beyond the specific configurations evaluated. By providing a systematic framework for assessing RAG system performance across multiple dimensions, we enable practitioners to make more informed implementation decisions based on their specific requirements and constraints. The multidimensional evaluation approach, incorporating both quality metrics and efficiency considerations, offers a more holistic view of system performance than traditional evaluations focused solely on output quality.

Several promising directions for future research emerge from this work. First, exploring adaptive configuration approaches that dynamically adjust parameters based on query characteristics and system conditions could enhance both performance and efficiency. Second, investigating the impact of emerging embedding models and retrieval techniques on the identified trade-offs could reveal new optimization opportunities. Third, extending this evaluation framework to domain-specific applications would provide more targeted guidance for practitioners in specialized fields.

Additionally, developing more sophisticated caching and preprocessing strategies could help mitigate the computational overhead of high-quality configurations, making them more viable for latency-sensitive applications. Research into hybrid approaches that combine the strengths of different vector stores, chunking strategies, and retrieval mechanisms could also yield more balanced performance profiles than the individual components evaluated in this study.

In conclusion, this research contributes to a more nuanced understanding of RAG system optimization that balances quality, efficiency, and reliability considerations. By illuminating the complex interplay between different configuration parameters and their combined effects on system outcomes, we provide practical insights for implementing effective RAG systems across diverse application domains. As these systems continue to evolve and find new applications, the optimization principles and evaluation methodologies presented in this work will remain valuable guides for both researchers and practitioners in the field.

\section{Acknowledgments}
The authors would like to thank Prince Sultan University and Alfaisal University for their support.

\bibliographystyle{IEEEtran}
\bibliography{biblio}

\end{document}